%% file: paper.tex
\documentclass[runningheads]{llncs}
\usepackage[T1]{fontenc}
\usepackage{graphicx}
\usepackage{booktabs}
\usepackage[misc]{ifsym}
\newcommand{\corr}{(\Letter)}
\usepackage{algorithm}
\usepackage{algpseudocode}

\usepackage{array}
\newcolumntype{?}{!{\vrule width 1pt}}

\usepackage[hidelinks]{hyperref}

\input{tex/math_commands}

\input{tex/custom_commands}

\usepackage{amsmath,amsfonts,bm,booktabs,multirow,pifont}

\begin{document}

\title{Generative Example-Based Explanations:\\Bridging the Gap between Generative Modeling and Explainability}
\tocauthor{Philipp Vaeth, Alexander M. Fruehwald, Benjamin Paassen, Magda Gregorova} 
\toctitle{Generative Example-Based Explanations:\\Bridging the Gap between Generative Modeling and Explainability}

\titlerunning{Generative Example-Based Explanations} 



\author{
Philipp Vaeth\inst{1,2}\orcidID{0000-0002-8247-7907}\corr
\and \\
Alexander M. Fruehwald\inst{1} 
\and \\
Benjamin~Paassen \inst{2}\orcidID{0000-0002-3899-2450}
\and \\
Magda Gregorová\inst{1}\orcidID{0000-0002-1285-8130}}

\authorrunning{Vaeth et al.}
\institute{Center for Artificial Intelligence and Robotics, Technical University of Applied Sciences Würzburg-Schweinfurt, Franz-Horn-Straße 2,
Würzburg, Germany\\
\email{\{philipp.vaeth,magda.gregorova\}@thws.de, kumardibyanshu05@gmail.com} \and 
Bielefeld University, Universitätsstraße 25, Bielefeld, Germany\\
\email{bpaassen@techfak.uni-bielefeld.de}}

\maketitle 

\begin{abstract}
Recently, several methods have leveraged deep generative modeling to produce example-based explanations of image classifiers. 
Despite producing visually stunning results, these methods are largely disconnected from classical explainability literature.
This conceptual and communication gap leads to misunderstandings and misalignments in goals and expectations. 
In this paper, we bridge this gap by proposing a probabilistic framework for example-based explanations, formally defining the example-based explanations in a probabilistic manner amenable for modelling via deep generative models while coherent with the critical characteristics and desiderata widely accepted in the explainability community. 
Our aim is on one hand to provide a constructive framework for the development of well-grounded generative algorithms for example-based explanations and, on the other, to facilitate communication between the generative and explainability research communities, foster rigor and transparency, and improve the quality of peer discussion and research progress in this promising direction.

 \\ 

\textbf{Reproducibility:} The code, the trained model weights and the supplementary material to reproduce the results is available at \url{https://github.com/philippvaeth/XAIDIFF}.
\end{abstract}



\section{Introduction}
As more AI applications get introduced into our daily lives, the need for trust, safety and ethical use of AI in these automated decision systems increases~\cite{goodman2017european,chatila2021trustworthy}.
Explainable Artificial Intelligence (XAI) is a research field that addresses these concerns of trust, transparency and fairness by supplementing decisions with explanations~\cite{gunning2017explainable}.

A well-known approach focusing on explanations of an algorithm decision for a specific input data point is that of counterfactual explanations~\cite{xaisurvey_3}.
These are examples that change some of the input features of the original data point in order to flip the decision of the algorithm to the desired class.
Unlike conceptually similar adversarial examples~\cite{goodfellow2014explaining}, the changes in the counterfactual examples aim to be perceptible and understandable by a human observer.
A classical example of a counterfactual from~\cite{counterfactuals}: \textit{``You were denied a loan because your annual income was £30,000. If your income had been £45,000, you would have been offered a loan.''}

Though well-developed for tabular data~\cite{sahakyanExplainableArtificialIntelligence2021}, transferring this concept to high-dimensional data such as images is not straightforward. 
Without careful control, introducing changes in the original pixel space may distort the image, rendering it unrealistic and hence difficult to interpret by the user.
These considerations have lead to the development of new techniques for producing counterfactual examples based on modern deep generative modeling~\cite{dvce,rw_diff1,ebcf}.

While these often showcase examples of stunning visual quality, the value of these methods is greatly diminished by their disconnect to the classical explainability desiderata and lack of rigorous, transparent and comparable quantitative evaluation. 

Motivated by this lack of systematic grounding of counterfactual methods based on deep generative modelling, \textbf{we propose a probabilistic framework for example-based explanations} formally defining counterfactual examples and their properties in a probabilistic manner amenable for modeling via deep generative models. 
This framework shall help in bridging the fields of generative modeling and explainability by aligning the terminology, the expectations, and by enabling a comparative quantitative evaluation of the methods fostering further scientific advancements. 

\begin{figure}[h]\label{local_xai}
 \centering
 \includegraphics[width=0.8\linewidth]{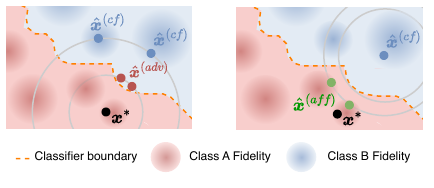}
 \caption{Relationships between example-based explanations. \emph{Left:} Original example $\ovx$ classified as A has high-fidelity - is in the high-density (dark) area of the data distribution. Adversarial examples $\hvx^{(adv)}$ are low-fidelity - in the low-density (light) area - near the original but classified as B. Counterfactuals $\hvx^{(cf)}$ are also classified as B, but they are high-fidelity (high density area). They are close to the original sample but may be further away than the adversarial examples. \emph{Right:} Affirmatives are high-fidelity samples classified as A near the counterfactuals. In consequence, they come from high-density areas within A which may or may not be close to $\ovx$.}
\end{figure}

In the following, we first introduce the proposed probabilistic framework in section \ref{sec:explanatoryExamples}.
We define the counterfactual examples, contrast them to adversarial examples and introduce a new complementary concept of affirmative examples in section \ref{sec:probDefs}.
Based on the introduced definitions, in section \ref{sec:GenerativeExplainerProblem} we formulate the search for counterfactual examples as a generative explainer problem, an optimization problem that serves as an objective for the deep generative approaches for counterfactuals.
Finally, in section \ref{sec:evaluation} we outline an evaluation procedure using the above definitions and propose concrete metrics usable in practice. 

To illustrate the applicability of the newly proposed framework for developing and evaluating new algorithms in practice, we describe in section \ref{sec:DDPM} an algorithm for counterfactual generations based on the classifier guidance~\cite{diffusionbeatgans} of denoising diffusion probabilistic models~\cite{ddpm}.
In addition to being well-grounded within the proposed probabilistic framework, the algorithm also addresses some shortcomings of previously proposed approaches for diffusion-based counterfactuals~\cite{dvce,rw_diff1}.
Through experiments in section \ref{sec:Experiments} we verify that our proposed framework and the generative explainer problem can indeed serve as a set of guiding principles in developing new generative algorithms for counterfactuals and that the evaluation procedure provides quantitative results that help in assessing the quality of the algorithm and produced counterfactual examples.
Here we also put forward and release a new synthetic SportBalls dataset (section~\ref{sec:Experiments_sportballs}) for a controlled explanation environment with varying complexity and intuitively unambiguous explanations.
We conclude our paper in section~\ref{sec:Conclusions} reflecting on some limitations of our framework and counterfactual explanations in general.

\section{Example-based explanations}\label{sec:explanatoryExamples}

Example-based explanations are well established in the local explainability domain, with counterfactual explanations as the most prominent type~\cite{xaisurvey_1,xaisurvey_2}.
The purpose of these is to help the user understand the decisions of an algorithm for a specific data example (local explanation).
By applying a minimal change to the original example which changes the algorithmic decision, the user shall understand which crucial changes separate the original example from the other class according to the classifier.

The foundational counterfactual explanation system was proposed in~\cite{counterfactuals}.
For an original data point $\ovx$ labeled by a classification algorithm $f_\theta$ as $y_o$ (`o' - original), a counterfactual explanation is a point $\hvx$ close to $\ovx$ with some of the input features altered so that it is labeled by $f_\theta$ as $\yt$ (`$\tau$' - target).
According to~\cite{counterfactuals}, the counterfactual $\hvx$ can be found by any suitable optimizer through solving the following minimization problem:
\begin{equation}\label{eq:wachter_explainer}
\min_{\hvx}
\lambda\left( f_{\theta}(\hvx) - \yt\right)^2
 +d\left(\ovx, \hvx\right)\enspace ,
\end{equation}
with $\lambda$ a hyperparameter trading off the label accuracy with a problem-specific distance $d$ between the original and the counterfactual example.

An important caveat to this seemingly simple problem 
is that of adversarial examples~\cite{goodfellow2014explaining}.
These are also examples that are close to the original $\ovx$ and flip the decision of the classifier $f_\theta$.
Formally, they can be obtained by solving the same minimization problem \eqref{eq:wachter_explainer}. 
However, a crucial difference between adversarial and counterfactual examples is that of human understandability. 
While adversarial examples fool the classifier by introducing possibly many tiny changes imperceptible by humans, the purpose of counterfactual examples is to provide a human with an understandable and hence perceptible explanation through sparse meaningful changes.

As~\cite{counterfactuals} notes, the critical difference is whether the example comes from a \emph{``possible world''}.
The imperceptible changes introduced by adversarial attacks can push the examples out of the data distribution which causes the classifier trained on-distribution to be easily fooled and make incorrect decisions. 
In contrast, what brings explanatory value to the counterfactual examples is that these are realistic in-distribution examples with changes observable and interpretable by a human observer.

Recognizing this challenge, the authors of~\cite{counterfactuals} decide to focus on simpler tabular datasets with directly interpretable input features and defer the problem to future research stating:
\textit{``Further research into how data from high-dimensional and highly-structured spaces, such as natural images, can be characterised is needed before counterfactuals can be reliably used as explanations in these spaces.''}

Recently emerging algorithms for counterfactual examples in high-dimensional image spaces based on deep generative modeling~\cite{dvce,rw_diff1,ebcf} are, despite their ability to produce visually stunning examples, largely ad-hoc and lack the connection to the initial and generally accepted counterfactual framework as put forward by~\cite{counterfactuals}.
Together with the lack of systematic and comparable evaluation of the methods, this leads to low trust in the methods from the XAI expert community partly aggravated by possible misunderstandings due to misaligned field-specific terminology.

This is where our paper comes in.
We propose a \textbf{probabilistic framework for example-based explanations} (section \ref{sec:probDefs}) that formally defines counterfactual, adversarial and newly introduced affirmative examples and their properties in a probabilistic manner.
This framework aligns with both the initial counterfactual explanation concepts as outlined in \cite{counterfactuals} and the probabilistic nature of deep generative model formulations.
As such it allows for integrating deep generative models into the process of searching for counterfactual examples through the optimization problem \eqref{eq:wachter_explainer} by a slight modification into a generative explainer problem defined in section \ref{sec:GenerativeExplainerProblem}.
It also enables systematic quantitative evaluation of the counterfactual algorithms and the generated examples (section \ref{sec:evaluation}), fostering trust through transparency and comparability.

\subsection{Probabilistic framework for example-based explanations}\label{sec:probDefs}

In view of the above discussion on the critical differences between counterfactual and adversarial examples and in line with the intuitive argumentation that an example is useful as an explanation only if it comes from a ``possible world'', we propose to use the \textbf{fidelity} of an example to the underlying data distribution as one of the central concepts of our probabilistic framework.
We opt to use fidelity as opposed to previously introduced and broader terms plausibility \cite{kenny2021Plausible} or realism to capture the specific property of the example to be in-distribution as measured by the data likelihood.

The other two concepts used in the proposed framework originate directly from the counterfactual definitions as introduced in \cite{counterfactuals} and equation~\eqref{eq:wachter_explainer}. 
The first is that of \textbf{closeness} measured by the distance $d$ between the original $\ovx$ and generated example $\hvx$, the second is the \textbf{validity} reflecting the correctness of the decision of the classification algorithm with respect to the target class. 


Let $\ovx \in \gX$ be a random sample from the true underlying generating data distribution $\pdata$ with the original classifier label $y_{o} \in \gY$;
$\hvx \in \gX$ a synthetically generated sample; 
$\yt \in \gY$ the target class label; 
$d: \gX \times \gX \to \R_+$ a suitable user-specified distance function (such as $\ell_1$, $\ell_2$ or some suitable perception distance).
We define the following concepts with respect to the probabilistic classification algorithm $\pcl$ and the original example $\ovx$:

\begin{definition}[Counterfactual explanation]
 We call a synthetic data sample $\hvx$ with \textbf{all} the following characteristics a targeted counterfactual explanation with respect to some target $\yt \in \gY$ and some thresholds $\delta$, $\epsilon > 0$:
 \begin{enumerate}
 \item $\hvx$ is \textbf{close} to the original sample $\ovx$: $d(\hvx, \ovx) \leq \delta$;
 \item $\hvx$ is \textbf{valid} w.r.t. the target label classification area: \\
 $\pcl(\rvY = \yt \mid \rvX = \hvx) > \pcl(\rvY = y_j \mid \rvX = \hvx), \text{ for all } j \neq \tau$ ;
 \item $\hvx$ has \textbf{high fidelity} to the real data distribution (is a realistic data example):
 $\pdata(\rvX = \hvx) \geq \epsilon$\enspace.
 \end{enumerate}
\end{definition}

\begin{definition}[Adversarial example]
 We call a synthetic data sample $\hvx$ with \textbf{all} the following characteristics a targeted adversarial example with respect to some target $\yt \in \gY$ and some thresholds $\delta$, $\epsilon > 0$:
 \begin{enumerate}
 \item $\hvx$ is \textbf{close} to the original sample $\ovx$: $d(\hvx, \ovx) \leq \delta$;
 \item $\hvx$ is \textbf{valid} w.r.t. the target label classification area: \\
 $\pcl(\rvY = \yt \mid \rvX = \hvx) > \pcl(\rvY = y_j \mid \rvX = \hvx), \text{ for all } j \neq \tau$ ;
 \item $\hvx$ has \textbf{low fidelity} to the real data distribution (is not a realistic data example):
 $\pdata(\rvX = \hvx) \leq \epsilon$\enspace.
 \end{enumerate}

 We point out that the only difference in the definitions of adversarial and counterfactual examples is the fidelity criterion which shall be high for counterfactuals and low for adversarial examples.
\end{definition}

Furthermore, we propose a new type of example-based explanations we call \textbf{affirmative examples}.
Conceptually, these can be thought of as counterfactual examples to counterfactuals, see the right part of figure~1.
Affirmatives are high-fidelity examples which re-introduce into the counterfactuals the critical features to bring them back to the original class.
They shall thus re-affirm the user understanding of the effects the feature changes have on the algorithm decisions.

A somewhat related yet different approach to local explainability which shall not be confused with our affirmatives are the contrastive explanations ~\cite{dhurandhar2018explanations}, which highlight features in the original example that need to be present or absent for the classifier to maintain its decision.
In contrast, affirmatives are new examples different from $\ovx$ that alter the counterfactual examples to bring them back to the same class as the original.

\begin{definition}[Affirmative examples]
 We call a synthetic data sample $\hvx$ with \textbf{all} the following characteristics an affirmative of a counterfactual $\hvx^{(cf)}$ with some thresholds $\delta$, $\epsilon > 0$:
 \begin{enumerate}
 \item $\hvx$ is \textbf{close} to the counterfactual explanation $\hvx^{(cf)}$:
 $d(\hvx, \hvx^{(cf)}) \leq \delta$ ;
 \item $\hvx$ is \textbf{valid} w.r.t. the original label $y_{o}$ classification area: \\
 $\pcl(\rvY = y_{o} \mid \rvX = \hvx) > \pcl(\rvY = y_j \mid \rvX = \hvx), \text{ for all } j \neq o$;
 \item $\hvx$ has \textbf{high fidelity} to the real data distribution (is a realistic data example):
 $\pdata(\rvX = \hvx) \geq \epsilon$\enspace.
 \end{enumerate}
\end{definition}

It is important to note that all the above concepts are local (i.e., pertaining to the specific original sample $\ovx$) and classifier-specific.
Furthermore, we can bring the above definitions into a fully probabilistic setting by reformulating the closeness constraints in the form of a prior distribution $p(\hvx|\ovx)$ on the data sample $\hvx$ (e.g., Gaussian for the $\ell_2$ or Laplace for the $\ell_1$ distances).

\subsection{Generative explainer problem}\label{sec:GenerativeExplainerProblem}
Following the definitions in section~\ref{sec:probDefs}, we formulate the search for example-based explanations as an optimization problem.
We give it here for the case of counterfactual examples; affirmatives and adversarial examples follow trivially in analogy.
\begin{equation*}
\begin{split}
 \argmax_{\hvx} \;\; & \pcl(\rvY = \yt \mid \rvX = \hvx) \hspace{1.85cm} \text{\small(validity)}\\
\text{s.t. } & 
\begin{aligned}[t]
 \pdata(\rvX = \hvx) & \geq \epsilon, & \hspace{1.6cm} \text{\small(fidelity)}\\
 d(\hvx, \ovx) &\leq \delta & \hspace{1.6cm} \text{\small(closeness)}\\
\end{aligned} 
\end{split}
\end{equation*}
\normalsize
In practice, we typically have no access to the true underlying data distribution $\pdata$ other than through the observed data and therefore cannot evaluate the fidelity of the point $\hvx$.
We propose to replace the unknown true distribution by an approximation $\pt \approx \pdata$ learned by a suitable generative algorithm, enabling us to sample high fidelity examples $\hvx \sim \pt$.
Our final formulation of the \textbf{generative explainer problem} reformulates the optimization into a Lagrangian form and subsumes the fidelity constraint into the sampling procedure:
\begin{align}\label{eq:gen_explainer}
 \underset{\hvx\sim\pt}{\operatorname{arg\, max}}\; &\lambda\, \pcl(\rvY =\yt \mid \rvX = \hvx) - d(\hvx, \ovx) \enspace .
\end{align} 
Though building up from our probabilistic definitions, this problem is consistent with the counterfactual explanation problem~\eqref{eq:wachter_explainer} with the additional fidelity constraint ensured implicitly through sampling from $\pt$.
We thus provide a bridge between our probabilistic framework for example-based explanations and the well established XAI concepts of counterfactual examples introduced in~\cite{counterfactuals}. 


Similarly to \eqref{eq:wachter_explainer}, the particular choice of the hyperparameter $\lambda$ and the distance function, depends on one hand on the properties of the data and the decision algorithm, and on the other on user preferences and interests. 

\subsection{Evaluation of example-based explanations}\label{sec:evaluation}
The evaluation of explanatory algorithms and of the quality of the explanations remains a challenge for both traditional tabular and high-dimensional image data \cite{Vaeth2023}.
Though human studies based on visual inspection provide valuable feedback, they are challenging to undertake and often suffer from a number of deficiencies due to practical considerations (e.g., expertise mismatch, motivation bias, experience learning curve, time constraints, etc.~\cite{lage2019evaluation,rosenfeld2021better}).

We acknowledge the usefulness and importance of such studies yet argue that quantitative evaluation allowing for comparability of obtained results can foster transparency and trust in the algorithms and serve as the first indicator of the quality of the explanations before much more lengthy and expensive human studies are undertaken.

For these reasons we propose a quantitative evaluation scheme for example-based explanations directly motivated by the probabilistic framework introduced in section \ref{sec:probDefs} and using the three characteristics therein: closeness, validity, and fidelity.
For illustration, we put the scheme to use in evaluating the explanations generated through our experiments in section \ref{sec:Experiments}.

To measure closeness of the examples we propose to use a suitable distance metric $d$, preferably the one chosen for the generative explainer problem \eqref{eq:gen_explainer}. 
Commonly used metrics include the $\ell_p$ distances, such as the $\ell_1$ distance for sparser changes or the $\ell_2$ distance for broader changes across the input. 


For validity, we propose to report the classifier prediction confidence for the target class.
Though the output of the neural network classifier may not be a reliable indicator for the classifier confidence~\cite{hein2019relu}, we consider it a good-enough proxy when comparing a single algorithm across multiple explanatory examples.

In the generative modelling community it is common to measure the alignment of generated examples with the underlying data distribution though metrics based on distribution matching such as the Fréchet inception distance~\cite{fid}, the Inception Score~\cite{is} or the Precision/Recall~\cite{precrecall}.
However, these measures are not suitable for evaluating the fidelity of an individual example.
We therefore propose to measure the fidelity of the explanatory example by its negative log-likelihood or its standard proxies, e.g., the variational lower bound relative to that of the original example $\ovx$. 
Interpreted as the bits necessary for lossless compression via an entropy encoding scheme optimized for the learned $\pt$, an explanatory example should require similar amount of bits as the original $\ovx$, whereas the encoding length of an adversarial should be much longer. 
To make this connection explicit, we present the fidelity in our experimental evaluation in section \ref{sec:Experiments_sportballs} as the additional bits-per-dim (bpd) required to encode the examples in comparison to the original data point.
Since the explanatory example may require fewer bits, this metric can be negative.

Though the primary purpose of the above metrics is the evaluation of the quality of the algorithm and the generated explanatory examples, they can also serve the user as a complementary explanatory information. 
For example, the user may generate a diverse set of explanatory examples by varying the strength of the hyperparameter $\lambda$ trading off the validity with the closeness of the examples in the generative explainer problem \eqref{eq:gen_explainer}. 
The metrics for the produced explanatory examples shall vary accordingly and thus further reinforce the user understanding of the classifier decision.
We illustrate this in our experiments in table \ref{tab:xaieval_images} and the related discussion.


\section{Diffusion-generated explanatory examples}\label{sec:DDPM}

In this section we showcase the use of the probabilistic framework introduced in section \ref{sec:explanatoryExamples} in a constructive manner through developing an algorithm for generative example-based explanations (XAIDIFF).
We choose to work with the class of Denoising Diffusion Probabilistic Models (DDPM)~\cite{ddpm} as the state of the art generative model for learning the underlying data distribution and sampling high fidelity examples, combined with the classifier guidance mechanism \cite{diffusionbeatgans} to steer the generations towards desirable output in terms of validity and closeness as per the definitions in section~\ref{sec:probDefs}.

Similar approaches have previously been proposed in \cite{dvce,rw_diff1}. 
However, the solutions these introduce are rather ad-hoc and mostly based on intuition rather than clear definitions linking back to the initial concepts of counterfactual explanations as discussed in section \ref{sec:explanatoryExamples}.
They also do not provide a systematic quantitative evaluation of their results that would allow for comparisons across methods.

Our XAIDIFF algorithm does not seek to outperform the above methods. 
Indeed, direct comparison is impossible due to the missing evaluation in the original papers and difficulties (despite our best efforts) in reproducing the original results.
Instead, our aim is to demonstrate how a generative approach to explanatory examples can be grounded within our probabilistic framework aligned with the classical counterfactual desiderate of \cite{counterfactuals}, and the advantages this brings for the subsequent quantitative evaluation.
In doing so, we also address some conceptual shortcomings of the previous algorithms.

\subsection{DDPM preliminaries}\label{sec:DDPMprelim}

DDPMs are latent variable models trained to approximate the underlying data distribution $\pdata$ via a model $\pt(\vx_0) = \int \pt(\vx_0 | \vx_1) \prod_{t=1}^T \pt(\vx_{t-1} | \vx_t) d \vx_{1:T}$.
They follow an encoder-decoder architecture:
the encoder (forward process) is defined as a Markov chain Gaussian model $q(\vx_t | \vx_{t-1}) :=\mathcal{N}\left(\vx_t ; \sqrt{1-\beta_t} \vx_{t-1}, \beta_t \mathbf{I}\right)$, progressively adding random Gaussian noise to an original example $\vx_0$ with a pre-defined variance schedule $\beta_1, \ldots, \beta_T$ over the diffusion steps $\vx_1, \ldots, \vx_T$ so that $\vx_T \sim \normaldist$;
the Gaussian Markov decoder (reverse process) 
\begin{equation}\label{eq:ddpmReverse}
 \pt(\vx_{t-1} \mid \vx_t) = \mathcal{N}\left(\vx_{t-1}; \decodermuthetaxt, \decodervariancefunctionshort\right) 
\end{equation}
with learned mean $\decodermuthetaxt$ and typically fixed diagonal covariance (a function of the schedule $\beta_t$)
reverts the forward diffusion process (encoder) by gradually denoising the signal $\vx_t$ to produce samples matching the original data distribution $\pt \approx \pdata$.

Instead of maximizing the intractable likelihood $\pt$ or its variational lower bound (ELBO), DDPMs can be trained by minimizing a simplified loss function matching the noise predicted by the denoising reverse process $\noiselearned$ to the ground truth noise $\noisegt$ of the forward process over time-steps $t$ sampled from a uniform distribution $\mathcal{U}(1, T)$:
\begin{equation*} \label{eq:ddpmsimpleloss}
 L:=\mathbb{E}_{t \sim\mathcal{U}(1, T), \vx_0 \sim q\left(\vx_0\right), \noisegt \sim \normaldist} \left\|\noisegt-\noiselearned\right\|^2 , \ \vx_t =\sqrt{\bar{\alpha}_t} \vx_0+\noisegt \sqrt{1-\bar{\alpha}_t} ,
\end{equation*} 
with $\bar{\alpha}_t=\prod_{s=1}^t (1-\beta_s)$.
For more details we encourage the interested reader to consult the abundant literature on DDPMs and related models, e.g., the comprehensive survey \cite{yangDiffusionModelsComprehensive2023}.

\subsection{Classifier guidance}\label{sec:ClasssifierGuidance}
The vanilla DDPM introduced in section~\ref{sec:DDPMprelim} enables synthesizing random samples $\hvx \sim \pt \approx \pdata$ with high fidelity to the underlying data distribution.
However, to generate valid class-specific data points
the reverse process shall in principle be conditioned on the target label $p_{\theta}\left(\vx_{t-1} \mid \vx_t, \yt\right)$.

Instead, the classifier guidance mechanism \cite{diffusionbeatgans} leverages an unconditionally trained (label-unaware) DDPM with unconditional denoising transitions
\eqref{eq:ddpmReverse}.
The guidance is achieved through shifting the mean $\decodermuthetaxt$ of the transitions by the gradients of a classifier $\nabla_{\vx_t} \log \pcl\left(\yt \mid \vx_t\right)$ trained over the noisy data $\vx_t$ produced by the forward process of the DDPM to yield class-conditional samples: 
\begin{equation}\label{eq:ClassifierGuidance}
 \begin{aligned}
 \decodermuthetaxtprime = \decodermuthetaxt
 + s \; \decodervariancefunctionshort \; \nabla_{\vx_t} \log \pcl\left(\yt \mid \vx_t\right) \enspace , 
 \end{aligned}
\end{equation}
where $s$ is a scaling factor controlling the strength of the classifier guidance.

\subsection{Classifier guidance for explanatory examples}\label{sec:challenges}
The mechanism for classifier guidance outlined in section~\ref{sec:ClasssifierGuidance} opens a window of opportunity for generating explanatory examples coherent with our probabilistic framework and the definitions in section~\ref{sec:probDefs}.
In particular, two of the critical characteristics for the explanatory examples can be expected to be satisfied by construction. 
The use of DDPM guarantees the high fidelity of the generated explanatory examples and the classifier guidance ensures their validity.
The approach, however, does not cater for the closeness of the examples, which is the third critical characteristics of our framework.

In accordance with the generative explainer problem \eqref{eq:gen_explainer}, we introduce a distance function into the classifier guidance formulation \eqref{eq:ClassifierGuidance} as follows: 
\begin{equation}\label{eq:guidance}
 \decodermuthetaxtprime = \decodermuthetaxt + \decodervariancefunctionshort \; \nabla_{\vx_t} \left[ \lambda \log \pcl\left(y \mid \xzeropred\right) - \norm{\xzeropred - \ovx}_1 \right] \enspace.
\end{equation}
Similar to \cite{dvce,rw_diff1} we chose the $\ell_1$ distance suggested as a \emph{``sensible first choice''} in \cite{counterfactuals} promoting sparsity in the introduced changes, and we replace the noisy DDPM samples $\vx_t$ with the DDPM one-step predictions of the denoised data $\xzeropred$ which are closer to the true data distribution than the noisy samples $\vx_t$.

However, in contrast to the previous methods \cite{dvce,rw_diff1}, we do not truncate the reverse diffusion sampling to start from the middle of the process but instead start from the very end, from the complete white noise $\vx_T \sim \mathcal{N(\mathbf{0}, \mathbf{I})}$. 
In this way, we do not constrain the search space for \eqref{eq:gen_explainer} and allow for higher diversity in the generated explanatory examples.
Furthermore, we propose to tackle the previously noticed issue of the instability of the classifier gradients head-on.
Instead of relying on a surrogate robust classifier to help in the guidance as in \cite{dvce},
we interpret the gradient guidance in equation \eqref{eq:guidance} as stochastic gradient descent steps.
Motivated by this view, we propose to leverage modern methods for stochastic optimization of non-stationary objectives with noisy gradients.
Concretely, we introduce ADAM~\cite{adam} into the guidance steps to automatically adapt the step size from estimates of the first and second moments of the gradients.

Finally, we link the guidance steps in equation \eqref{eq:guidance} back to the definitions in section~\ref{sec:explanatoryExamples} and interpret the individual terms as gradients of the generative explainer problem \eqref{eq:gen_explainer} introduced in section~\ref{sec:GenerativeExplainerProblem}:
\begin{itemize}
 \item $\decodermeanlearnedshort$ implicitly as the gradient towards maximum fidelity, 
 \item $\nabla_{\vx_t}\big[\log \pcl\left(y \mid \xzeropred\right)\big]$ as the gradient towards maximum validity, and 
 \item $\nabla_{\vx_t}\big[\norm{\xzeropred - \ovx}_1\big]$ as the gradient towards maximum closeness.
\end{itemize}

\section{Experiments}\label{sec:Experiments}

In this section, we put our XAIDIFF algorithm into use and evaluate its ability to generate explanatory examples showcasing the constructive value of our probabilistic framework.

%

\subsection{Datasets \& experimental setup}
We provide experimental results on two main data sets.
The first is a newly proposed synthetic data set, named SportBalls, carefully designed to illustrate they key components of our framework by providing a controlled explanation scenario where the optimal explanation is intuitively clear.
The second is 
the real-world data set CelebA~\cite{celeba}.
Additional results for a third (MNIST~\cite{mnist}) are provided in the appendix.

\emph{SportBalls} is a synthetically created dataset which contains multiple similar objects with slight variations in terms of scaling, shape, size, rotation and placement but clear semantic differences (i.e., colors and pattern).
Each image contains three out of five randomly selected sport balls placed at random coordinates on white background with random rotation and scaling.
There can be multiple types of sport balls in one image and the balls may overlap (even occlude each other), creating a more complex yet controlled explanation setting.
For the experiments, we pick four starting images each containing at least one soccer ball (the original class) with varying complexity for the explanation: the first image where the sport balls are non-overlapping, the second image with multiple soccer balls, the third image where the soccer ball is the background of another sport ball, and the fourth image where the soccer ball is partly occluded.
The code to generate the dataset is provided for further re-use by the community (see the links in the title page). 

\emph{CelebA} is a publicly available data set of celebrity faces with easily recognizable attributes for classification. 
We downscale the data to 64x64 pixels.
As starting points for explanations, we pick four images of faces with negative values for all three attributes: no eyeglasses, no smile, not blond.
These are not cherry-picked examples from experimentation but were selected a priori due to their diversity in terms of gender, hairstyle, formal attire and background.

\emph{Classifiers \& DDPM:}
In a realistic explainability scenario the classifier to be explained is provided externally. 
In our experiments we therefore pick a standard architecture for the classifiers, a low-resource small version of the MobileNetV3~\cite{howard2019searching} from `torchvision', and train a binary classifier (soccerball=yes/no) to 96.8 \% test accuracy over the SportBalls dataset, and a multi-label classifier eyeglasses=yes/no, smile=yes/no, blond\_hair=yes/no for CelebA to 97.13\%, 89.71\% and 93.33\% test accuracy respectively.
Our XAIDIFF algorithm assumes access to a DDPM trained over the same data distribution as the classifier.
We therefore train the DDPM over the same data sets.
The implementation relies on the open-source Diffusers toolbox~\cite{von-platen-etal-2022-diffusers}.

\emph{Evaluation:}
We evaluate closeness, validity and fidelity of the XAIDIFF generations according to the scheme in section~\ref{sec:evaluation}.
We measure the closeness by the $\ell_1$-distance in coherence with equation~\eqref{eq:guidance}.
For validity we report the classifier prediction confidence for the target class.
Finally, we use the variational lower bound (ELBO) of the learned model $\pt$ to calculate the resulting bits-per-dim and report the additional bits of the explanatory example as the measure of fidelity. 
We employ the above metrics to document the correct functioning of the XAIDIFF algorithm.
For this, we compare the averages of the metrics across sets of 16 explanatory examples generated by the algorithm. 
For instance, a well functioning algorithm shall ensure that counterfactual examples have on average higher fidelity than adversarial examples.

\subsection{Generative example-based explanations}\label{sec:Experiments_sportballs} 
In this section we validate that XAIDIFF produces example-based explanations as per our definitions in section \ref{sec:probDefs}.

\begin{table}[!ht]
 \centering 
 \def \imgwidth {0.089}
 \caption{Original SportBalls/CelebA examples $\ovx$ compared to XAIDIFF counterfactuals $\hvx^{(cf)}$, adversarial examples $\hvx^{(adv)}$, and generations without closeness $\hvx^{(cf\setminus c)}$ and validity criteria $\hvx^{(cf\setminus v)}$, respectively.}
 \label{tab:sportballs_cf_images}\label{tab:cf_images} 
 \begin{tabular}{@{}ccc|cc?ccc|cc@{}}
 \toprule
 \multicolumn{5}{c?}{\textbf{Sportballs}} & \multicolumn{5}{c}{\textbf{CelebA}} \\
 \midrule
 $\ovx$ & $\hvx^{(cf)}$ & $\hvx^{(adv)}$ & $\hvx^{(cf\setminus c)}$ & $\hvx^{(cf\setminus v)}$ & $\ovx$ & $\hvx^{(cf)}$ & $\hvx^{(adv)}$ & $\hvx^{(cf\setminus c)}$ & $\hvx^{(cf\setminus v)}$ \\ \midrule
 \centering\arraybackslash
 \includegraphics[width=\imgwidth\linewidth,trim={0.6cm 0 0 0},clip]{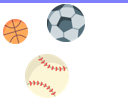} & \includegraphics[width=\imgwidth\linewidth,trim={0.6cm 0 0 0},clip]{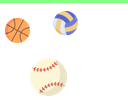} & \includegraphics[width=\imgwidth\linewidth,trim={0.6cm 0 0 0},clip]{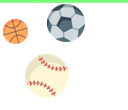} & \includegraphics[width=\imgwidth\linewidth,trim={0.6cm 0 0 0},clip]{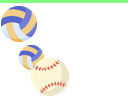} & \includegraphics[width=\imgwidth\linewidth,trim={0.6cm 0 0 0},clip]{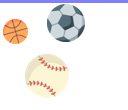} & \includegraphics[width=\imgwidth\linewidth]{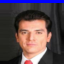} & \includegraphics[width=\imgwidth\linewidth]{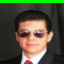} & \includegraphics[width=\imgwidth\linewidth]{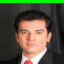} & \includegraphics[width=\imgwidth\linewidth]{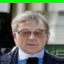} & \includegraphics[width=\imgwidth\linewidth]{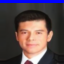} \\
 \includegraphics[width=\imgwidth\linewidth,trim={0.6cm 0 0 0},clip]{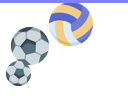} & \includegraphics[width=\imgwidth\linewidth,trim={0.6cm 0 0 0},clip]{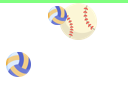} & \includegraphics[width=\imgwidth\linewidth,trim={0.6cm 0 0 0},clip]{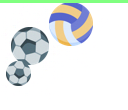} & \includegraphics[width=\imgwidth\linewidth,trim={0.6cm 0 0 0},clip]{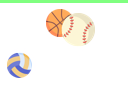} & \includegraphics[width=\imgwidth\linewidth,trim={0.6cm 0 0 0},clip]{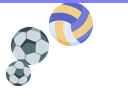} & \includegraphics[width=\imgwidth\linewidth]{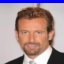} & \includegraphics[width=\imgwidth\linewidth]{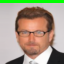} & \includegraphics[width=\imgwidth\linewidth]{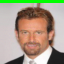} & \includegraphics[width=\imgwidth\linewidth]{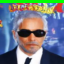} & \includegraphics[width=\imgwidth\linewidth]{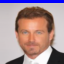} \\
 \includegraphics[width=\imgwidth\linewidth,trim={0.6cm 0 0 0},clip]{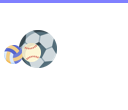} & \includegraphics[width=\imgwidth\linewidth,trim={0.6cm 0 0 0},clip]{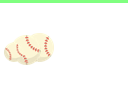} & \includegraphics[width=\imgwidth\linewidth,trim={0.6cm 0 0 0},clip]{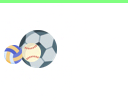} & \includegraphics[width=\imgwidth\linewidth,trim={0.6cm 0 0 0},clip]{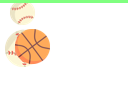} & \includegraphics[width=\imgwidth\linewidth,trim={0.6cm 0 0 0},clip]{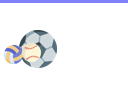} & \includegraphics[width=\imgwidth\linewidth]{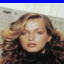} & \includegraphics[width=\imgwidth\linewidth]{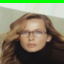} & \includegraphics[width=\imgwidth\linewidth]{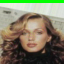} & \includegraphics[width=\imgwidth\linewidth]{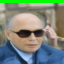} & \includegraphics[width=\imgwidth\linewidth]{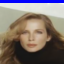} \\
 \includegraphics[width=\imgwidth\linewidth,trim={0.6cm 0 0 0},clip]{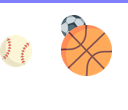} & \includegraphics[width=\imgwidth\linewidth,trim={0.6cm 0 0 0},clip]{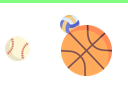} & \includegraphics[width=\imgwidth\linewidth,trim={0.6cm 0 0 0},clip]{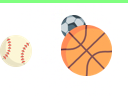} & \includegraphics[width=\imgwidth\linewidth,trim={0.6cm 0 0 0},clip]{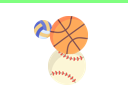} & \includegraphics[width=\imgwidth\linewidth,trim={0.6cm 0 0 0},clip]{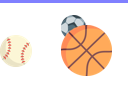} & \includegraphics[width=\imgwidth\linewidth]{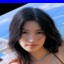} & \includegraphics[width=\imgwidth\linewidth]{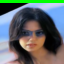} & \includegraphics[width=\imgwidth\linewidth]{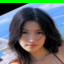} & \includegraphics[width=\imgwidth\linewidth]{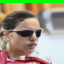} & \includegraphics[width=\imgwidth\linewidth]{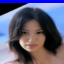} \\ 
 \bottomrule
 \end{tabular}
\end{table}

\begin{table}[!ht]
 \setlength{\tabcolsep}{2.5mm} 
 \renewcommand{\arraystretch}{1.25} 
 \centering
 \caption{Evaluation metrics for SportBalls/CelebA. \textbf{c}loseness: $\ell_1$-distance per dim, \textbf{v}alidity: classifier confidence (\%), \textbf{f}idelity: extra bpd compared to $\ovx$.}\label{tab:sportballs_cf_metrics}
 \begin{tabular}{@{}c|ccc|cc@{}}
 \toprule
 & & $\hvx^{(cf)}$ & $\hvx^{(adv)}$ & $\hvx^{(cf\setminus c)}$ & $\hvx^{(cf\setminus v)}$ \\ \midrule
 \multirow{3.3}{*}{\rotatebox{90}{\scriptsize \textbf{SportBalls}}} & \textbf{c}$_\downarrow$ & 0.7 (0.3) & 0.0 (0.0) & 1.2 (0.3) & 0.1 (0.1) \\
 & \textbf{v}$_\uparrow$ & 100 (0.0) & 83.4 (29.6) & 100 (0.0) & 51.5 (44.2) \\
 & \textbf{f}$_\downarrow$ & +0.3 (0.4) & +0.4 (0.1) & +0.6 (0.4) & 0.0 (0.1) \\ \midrule
 \multirow{3.3}{*}{\rotatebox{90}{\scriptsize \textbf{CelebA}}} & \textbf{c}$_\downarrow$ & 0.6 (0.1) & 0.1 (0.0) & 4.7 (0.7) & 0.6 (0.1) \\
 & \textbf{v}$_\uparrow$ & 99.5 (0.3) & 86.5 (20.7) & 99.3 (0.8) & 49.9 (49.1) \\
 & \textbf{f}$_\downarrow$ & -0.4 (0.2) & +1.1 (0.4) & +0.2 (0.4) & -0.5 (0.2) \\ \bottomrule
 \end{tabular}
\end{table}

In tables~\ref{tab:sportballs_cf_images} and \ref{tab:sportballs_cf_metrics} we contrast the counterfactual examples generated by XAIDIFF $\hvx^{(cf)}$ with adversarial examples $\hvx^{(adv)}$ obtained from the original examples through simple gradient updates $\hvx \leftarrow \hvx + \nabla_{\hvx} \, \pcl(\yt \mid \hvx)$ within the pixel space towards the classification target.
As an ablation of the XAIDIFF properties we remove the distance function in equation \eqref{eq:guidance} to obtain examples without the closeness criterion $\hvx^{(cf\setminus c)}$.
Finally, $\hvx^{(cf\setminus v)}$ disregard the validity criterion by removing the classifier term from equation \eqref{eq:guidance}.

We show one randomly selected generated example per starting image $\ovx$ in table~\ref{tab:sportballs_cf_images}.
More are shown in the appendix.
The metrics presented in table~\ref{tab:sportballs_cf_metrics} are averages (and standard deviations) calculated over sets of 16 examples.

Table~\ref{tab:sportballs_cf_images} illustrates that all starting images $\ovx$ of both datasets are correctly classified as soccerball=yes and eyeglasses=no (indicated by a blue stripe on top of the images).
All counterfactual explanations and adversarial examples flip the classifier decision (indicated by a green stripe).
However, only the counterfactuals correctly alter the original images. 
There are no soccer balls and the faces wear eyeglasses in the $\hvx^{(cf)}$ columns while the changes in the $\hvx^{(adv)}$ are imperceptible to human eye and therefore useless as explanations. 
Similarly, the class is changed and the images altered in the $\hvx^{(cf\setminus c)}$ column.
However, the images look very different than (not close to) the originals and therefore are of little explanatory value.
Examples in the $\hvx^{(cf\setminus v)}$ column are very close to the originals, however, they do not change the images sufficiently to flip the classifier decision.

The above visual analysis can be complemented by inspecting the numerical results in table \ref{tab:sportballs_cf_metrics}. 
While counterfactual examples $\hvx^{(cf)}$ are indeed closer to the originals than the $\hvx^{(cf\setminus c)}$ examples ignoring the closeness criterion, the adversarials $\hvx^{(adv)}$ are even closer though with lower fidelity. 
The high validity of all three corresponds to them successfully flipping the classifier decision, which fails for $\hvx^{(cf\setminus v)}$.
Yet, it is these examples disregarding the validity $\hvx^{(cf\setminus v)}$, which are the most fidel to the underlying data distribution as they do not need to balance the diffusion steps with the classifier gradient steps in equation \ref{eq:guidance}.

In tables~\ref{tab:aff_images} and~\ref{tab:aff_metrics}, we analyse the newly proposed affirmative explanations using the CelebA data set.
We contrast the affirmatives to the originals and the counterfactuals to demonstrate their explanatory value.
As we see in the tables, the affirmatives are high-fidelity examples classified the same as the original examples (blue stripe) and they indeed do not contain the concepts introduced by the counterfactual explanations. 
Though being close to the originals, they are even closer to the counterfactuals (see closeness in table~\ref{tab:aff_metrics}).
These characteristics are in line with the definitions provided in section~\ref{sec:probDefs} and experimentally demonstrate the intuitions depicted in the schematic illustration in figure~1.
\begin{table}[ht]
 \setlength{\tabcolsep}{1.25mm} 
 \centering
 \caption{\normalsize Affirmative examples $\hvx^{(aff)}$. Single-label for eyeglasses=no; multi-label for eyeglasses=no \underline{and} smiling=no \underline{and} blond=no.}\label{tab:aff_images}
 \def \imgwidth {0.13}
 \begin{tabular}{@{}c|ccc|ccc@{}}
 \toprule
 \multirow{11}{*}{\rotatebox{90}{\small\textbf{Single-label}}}
 & $\ovx$ & $\hvx^{(cf)}$ & $\hvx^{(aff)}$ &$\ovx$ & $\hvx^{(cf)}$ & $\hvx^{(aff)}$ \\ \midrule
 & \includegraphics[width=\imgwidth\linewidth]{media/images/aa9d0de3280e3932460c.png} & \includegraphics[width=\imgwidth\linewidth]{media/images/1b5d43ac679cd7beacc8.png} & \includegraphics[width=\imgwidth\linewidth]{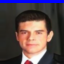} & \includegraphics[width=\imgwidth\linewidth]{media/images/f2402cb72cb4bb81f473.png} & \includegraphics[width=\imgwidth\linewidth]{media/images/708d5ec76142e735c3b3.png} & \includegraphics[width=\imgwidth\linewidth]{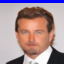} \\ 
 \cmidrule{2-7}
 & \includegraphics[width=\imgwidth\linewidth]{media/images/b237d5f7a3b0bedc6279.png} & \includegraphics[width=\imgwidth\linewidth]{media/images/f342595474104c53b0ba.png} & \includegraphics[width=\imgwidth\linewidth]{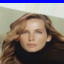} & \includegraphics[width=\imgwidth\linewidth]{media/images/8909cf31ce2b8f257ca0.png} & \includegraphics[width=\imgwidth\linewidth]{media/images/0e8e6a65a44a79e32a43.png} & \includegraphics[width=\imgwidth\linewidth]{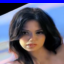} \\ 
 \midrule
 \multirow{2.6}{*}{\rotatebox{90}{\small\textbf{Multi-label}}}
 & \includegraphics[width=\imgwidth\linewidth]{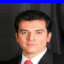} & \includegraphics[width=\imgwidth\linewidth]{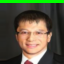} & \includegraphics[width=\imgwidth\linewidth]{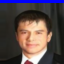} & \includegraphics[width=\imgwidth\linewidth]{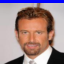} & \includegraphics[width=\imgwidth\linewidth]{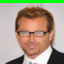} & \includegraphics[width=\imgwidth\linewidth]{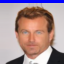} \\ 
 \cmidrule{2-7}
 & \includegraphics[width=\imgwidth\linewidth]{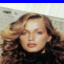} & \includegraphics[width=\imgwidth\linewidth]{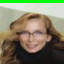} & \includegraphics[width=\imgwidth\linewidth]{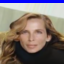} & \includegraphics[width=\imgwidth\linewidth]{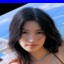} & \includegraphics[width=\imgwidth\linewidth]{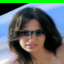} & \includegraphics[width=\imgwidth\linewidth]{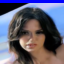} \\ 
 \bottomrule
 \end{tabular}
\end{table}
\begin{table}[ht]
 \setlength{\tabcolsep}{2.5mm} 
 \renewcommand{\arraystretch}{1.25} 
 \centering 
 \caption{\normalsize Evaluation metrics for affirmative examples.}\label{tab:aff_metrics}
 \begin{tabular}{@{}c|l|cc@{}|c|cc@{}}
 \toprule
 \multirow{7.2}{*}{\rotatebox{90}{\small \textbf{Single-label}}}
 & metrics & $\hvx^{(cf)}$ & $\hvx^{(aff)}$ 
 & & $\hvx^{(cf)}$ & $\hvx^{(aff)}$ \\
 \midrule
 &\textbf{c}$_\downarrow$ $(\ovx)$ & 0.6 (0.1) & 0.6 (0.1) 
 & \multirow{4.35}{*}{\rotatebox{90}{\small \textbf{Multi-label}}}
 & 0.57 (0.12) & 0.64 (0.15) \\
 &\textbf{c}$_\downarrow$ $(\hvx^{(cf)})$ & - & 0.2 (0.1) & & - & 0.24 (0.04) \\
 &\textbf{v}$_\uparrow$ & 99.5 (0.3) & 99.5 (0.4) & & 89.17 (4.95) & 97.24 (1.13) \\
 &\textbf{f}$_\downarrow$ & -0.1 (0.2) & -0.1 (0.2) & & -0.27 (0.18) & -0.25 (0.23) \\ 
 \bottomrule
 \end{tabular}
\end{table}

The above findings also
naturally extend to more complex multi-label scenarios: the XAIDIFF counterfactuals and affirmatives correctly trigger changes in the classifier decisions, flipping the classifications for all three attributes together - from no in the original to yes in the counterfactuals and back to no in the affirmatives.
All these changes are also corroborated by high validity values in table~\ref{tab:aff_metrics} and clearly observable in the generated images in table~\ref{tab:aff_images}. 

\begin{table}[!ht]
    \setlength{\tabcolsep}{2.5mm} 
    \renewcommand{\arraystretch}{1.25} 
    \def \imgwidth {0.15}
    \centering 
    \caption{Metrics as supplementary explanatory information - validity-closeness trade-off $\lambda \in \{0.1, 0.3, 0.5, 0.7\}$} \label{tab:xaieval_images} 
    \begin{tabular}{@{}c|cccc@{}}
    \toprule
    $\ovx$ & $\hvx^{(cf)}_{0.1}$ & $\hvx^{(cf)}_{0.3}$ & $\hvx^{(cf)}_{0.5}$ & $\hvx^{(cf)}_{0.7}$\\ \midrule
    \includegraphics[width=\imgwidth\linewidth]{media/images/3b75ad6333418ff4ba2f.png} & \includegraphics[width=\imgwidth\linewidth]{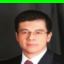} & \includegraphics[width=\imgwidth\linewidth]{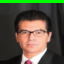} & \includegraphics[width=\imgwidth\linewidth]{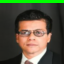} & \includegraphics[width=\imgwidth\linewidth]{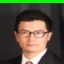}\\ \midrule
    \textbf{c}$_\downarrow$ & \textbf{0.37} & \textbf{0.37} & 0.49 & 0.50 \\
    \textbf{v}$_\uparrow$ & 96.23 &98.91 &\textbf{99.92} &99.74 \\
    \textbf{f}$_\downarrow$ & \textbf{-0.56} &-0.52 &-0.51 &\textbf{-0.56}\\
    \bottomrule
    \end{tabular}
\end{table}
As explained in section~\ref{sec:evaluation}, an additional benefit of the evaluation metrics is their explanatory value for the user.
We illustrate this in table~\ref{tab:xaieval_images}, where we supplement four counterfactual explanations with their respective metrics. The counterfactuals were generated with varying values of the validity-closeness trade-off hyperparameter $\lambda$ in equation \eqref{eq:guidance}.
Referring to the metric values, the user can better appreciate the difference between the counterfactual examples, which in turn shall help to explain the classifier decision. 
For instance, the validity of the first counterfactual in table~\ref{tab:xaieval_images} is lower than of the other three examples which corresponds to smaller recognizability of the eyeglasses in the image.
Giving the responsibility of selecting a good explanation to the user with very individual preferences may be an interesting direction for future work.
\section{Conclusions}\label{sec:Conclusions}
Recent methods for generative example-based explanations \cite{dvce,rw_diff1,ebcf} show impressive sample quality outlining a promising direction for future research.
However, these methods are largely inconsistent in used definitions, experimental baselines, approaches to evaluation and are largely disconnected from generally accepted XAI concepts often disregarding basic assumptions and expectations of the field.
This disconnect between the generative modeling and explainability communities leads to many misunderstandings.
To bridge this gap, this paper puts forward a probabilistic framework for example-base explanations consistent with traditional explainability desiderata and demonstrates how it can be used constructively for developing new generative-based explanatory algorithms by showcasing the diffusion-based XAIDIFF algorithm.
In line with the best research practice, we foster transparency and reproducibility by providing the complete implementation of our method together with the settings and the model files to reproduce the results of our experiments (see the links in the title page).

A major limitation of our method is in its generality.
Apart from small tweaks such as the $\lambda$ hyperparameter, the algorithm does not allow to individualize the model according to personal experience or affinity.
The metrics we propose are derived from our definitions and do not measure the appropriateness of the examples for improved understanding of a specific user.
Furthermore, our method assumes the availability of a diffusion model trained over the same (or sufficiently similar) data as the classifier. This may be impossible in some situations. 
There are also no mathematical guarantees for the proposed method.
Finally, we wish to warn that as any other generative model, our method could be used to produce malicious content.
This would, however, require access to a white-box classifier trained over malicious attributes which reduces the risk of misuse. 


%
%
%
\bibliographystyle{splncs04}

%
\end{document}

%% file: tex/math_commands.tex
\usepackage{amsmath}

\def\1{\bm{1}}

\def\vx{{\bm{x}}}

\DeclareMathAlphabet{\mathsfit}{\encodingdefault}{\sfdefault}{m}{sl}
\SetMathAlphabet{\mathsfit}{bold}{\encodingdefault}{\sfdefault}{bx}{n}

\def\gX{{\mathcal{X}}}
\def\gY{{\mathcal{Y}}}

\newcommand{\pdata}{p_{\rm{data}}}

\newcommand{\R}{\mathbb{R}}

\DeclareMathOperator*{\argmax}{arg\,max}

%% file: tex/custom_commands.tex
\usepackage{xcolor}
\usepackage{import}
\usepackage{subcaption} 

\usepackage{amsmath,amsfonts,bm,booktabs,multirow,pifont,algorithm,caption}

\usepackage[textsize=tiny]{todonotes}

\newcommand{\norm}[1]{\left\lVert#1\right\rVert}%
\newcommand{\ovx}{\bm{x}^*}

\newcommand{\yt}{y_\tau}
\newcommand{\hvx}{\bm{\hat{x}}}

\newcommand{\pcl}{p_{\rm{cl}}}
\newcommand{\pt}{p_{\boldsymbol{\theta}}}

\newcommand{\rvX}{\bm{X}}
\newcommand{\rvY}{\bm{Y}}

\newcommand{\xzeropred}{\hat{\vx}_0^{(\vx_t)}}

\newcommand{\decodermeanlearnedshort}{\mu_\theta}
\newcommand{\decodermuthetaxt}{\mu_\theta(\vx_t)}
\newcommand{\decodermuthetaxtprime}{\mu_\theta'(\vx_t)}

\newcommand{\decodervariancefunctionshort}{\Sigma(t)}
\newcommand{\noiselearned}{\hat{\boldsymbol{\epsilon}}_{\boldsymbol{\theta}}\left(\vx_t, t\right)}
\newcommand{\noisegt}{\boldsymbol{\epsilon}^*}
\newcommand{\normaldist}{\mathcal{N}(0, I)}

\def\pt{p_\theta}

%% file: paper.bbl
\begin{thebibliography}{10}
\providecommand{\url}[1]{\texttt{#1}}
\providecommand{\urlprefix}{URL }
\providecommand{\doi}[1]{https://doi.org/#1}

\bibitem{adam}
Diederik P. Kingma and Jimmy Ba: Adam: A Method for Stochastic Optimization. ICLR (2015)

\bibitem{artelt2020convex}
Artelt, Andr{\'e} and Hammer, Barbara: Convex Density Constraints for Computing Plausible Counterfactual Explanations. ICANN (2020)

\bibitem{bishopPatternRecognitionMachine2006}
Bishop, Christopher M.: Pattern Recognition and Machine Learning.  (2006)

\bibitem{blendeddiffusion}
Avrahami, Omri and Lischinski, Dani and Fried, Ohad: Blended Diffusion for Text-Driven Editing of Natural Images. CVPR (2022)

\bibitem{burda2015importance}
Yuri Burda and Roger Grosse and Ruslan Salakhutdinov: Importance Weighted Autoencoders. arXiv:1509.00519 (2016)

\bibitem{cchvae}
Pawelczyk, Martin and Broelemann, Klaus and Kasneci, Gjergji: Learning Model-Agnostic Counterfactual Explanations for Tabular Data. WWW (2020)

\bibitem{celeba}
Liu, Ziwei and Luo, Ping and Wang, Xiaogang and Tang, Xiaoou: Deep Learning Face Attributes in the Wild. ICCV (2015)

\bibitem{cem}
Dhurandhar, Amit and Chen, Pin-Yu and Luss, Ronny and Tu, Chun-Chen and Ting, Paishun and Shanmugam, Karthikeyan and Das, Payel: Explanations Based on the Missing: Towards Contrastive Explanations With Pertinent Negatives. NeurIPS (2018)

\bibitem{chatila2021trustworthy}
Chatila, Raja and Dignum, Virginia and Fisher, Michael and Giannotti, Fosca and Morik, Katharina and Russell, Stuart and Yeung, Karen: Trustworthy Ai. Reflections on Artificial Intelligence for Humanity (2021)

\bibitem{counterfactuals}
Wachter, Sandra and Mittelstadt, Brent and Russell, Chris: Counterfactual Explanations Without Opening the Black Box: Automated Decisions and the GDPR. JOLT (2017)

\bibitem{cruds}
Downs, Michael and Chu, Jonathan L and Yacoby, Yaniv and Doshi-Velez, Finale and Pan, Weiwei: Cruds: Counterfactual Recourse Using Disentangled Subspaces. ICML (2020)

\bibitem{ddpm}
Ho, Jonathan and Jain, Ajay and Abbeel, Pieter: Denoising Diffusion Probabilistic Models. NeurIPS (2020)

\bibitem{delaney2021uncertainty}
Eoin Delaney and Derek Greene and Mark T. Keane: Uncertainty Estimation and Out-of-Distribution Detection for Counterfactual Explanations: Pitfalls and Solutions. ICML Workshop on Algorithmic Recourse (2021)

\bibitem{dhurandhar2018explanations}
Dhurandhar, Amit and Chen, Pin-Yu and Luss, Ronny and Tu, Chun-Chen and Ting, Paishun and Shanmugam, Karthikeyan and Das, Payel: Explanations Based on the Missing: Towards Contrastive Explanations With Pertinent Negatives. NeurIPS (2018)

\bibitem{dice}
Mothilal, Ramaravind K and Sharma, Amit and Tan, Chenhao: Explaining Machine Learning Classifiers Through Diverse Counterfactual Explanations. ACM FAccT (2020)

\bibitem{dice2}
Kommiya Mothilal, Ramaravind and Mahajan, Divyat and Tan, Chenhao and Sharma, Amit: Towards Unifying Feature Attribution and Counterfactual Explanations: Different Means to the Same End. AAAI (2021)

\bibitem{diffusion_thermo}
Sohl-Dickstein, Jascha and Weiss, Eric and Maheswaranathan, Niru and Ganguli, Surya: Deep Unsupervised Learning Using Nonequilibrium Thermodynamics. ICML (2015)

\bibitem{diffusionbeatgans}
Dhariwal, Prafulla and Nichol, Alexander: Diffusion Models Beat Gans on Image Synthesis. NeurIPS (2021)

\bibitem{dinh2023pixelasparam}
Dinh, Anh-Dung and Liu, Daochang and Xu, Chang: PixelAsParam: A Gradient View on Diffusion Sampling With Guidance. ICML (2023)

\bibitem{dsprites}
Loic Matthey and Irina Higgins and Demis Hassabis and Alexander Lerchner: dSprites: Disentanglement Testing Sprites Dataset. Github (2017)

\bibitem{dvce}
Augustin, Maximilian and Boreiko, Valentyn and Croce, Francesco and Hein, Matthias: Diffusion Visual Counterfactual Explanations. NeurIPS (2022)

\bibitem{ebcf}
Guidotti, Riccardo: Counterfactual Explanations and How to Find Them: Literature Review and Benchmarking. KDD (2022)

\bibitem{emap}
Matt Chapman-Rounds and Marc-Andre Schulz and Erik Pazos and Konstantinos Georgatzis: EMAP: Explanation by Minimal Adversarial Perturbation. arXiv:1912.00872 (2019)

\bibitem{fid}
Heusel, Martin and Ramsauer, Hubert and Unterthiner, Thomas and Nessler, Bernhard and Hochreiter, Sepp: GANs Trained by a Two Time-Scale Update Rule Converge to a Local Nash Equilibrium. NeurIPS (2017)

\bibitem{fimap}
Chapman-Rounds, Matt and Bhatt, Umang and Pazos, Erik and Schulz, Marc-Andre and Georgatzis, Konstantinos: FIMAP: Feature Importance by Minimal Adversarial Perturbation. AAAI (2021)

\bibitem{goodfellow2014explaining}
Ian J. Goodfellow and Jonathon Shlens and Christian Szegedy: Explaining and Harnessing Adversarial Examples. ICLR (2015)

\bibitem{goodman2017european}
Goodman, Bryce and Flaxman, Seth: European Union Regulations on Algorithmic Decision-Making and a “Right to Explanation”. ICML WHI Workshop (2016)

\bibitem{gopfert2020adversarial}
G{\"o}pfert, Jan Philip and Artelt, Andr{\'e} and Wersing, Heiko and Hammer, Barbara: Adversarial Attacks Hidden in Plain Sight. IDA (2020)

\bibitem{gower2020variance}
Gower, Robert M and Schmidt, Mark and Bach, Francis and Richt{\'a}rik, Peter: Variance-Reduced Methods for Machine Learning. IEEE (2020)

\bibitem{gradcheck}
Philipp Vaeth and Alexander M. Fruehwald and Benjamin Paassen and Magda Gregorova: GradCheck: Analyzing Classifier Guidance Gradients for Conditional Diffusion Sampling. arXiv:2406.17399 (2024)

\bibitem{guidottiCounterfactualExplanationsHow2024}
Guidotti, Riccardo: Counterfactual Explanations and How to Find Them: Literature Review and Benchmarking. Data Min. Knowl. Discov. (2024)

\bibitem{gunning2017explainable}
Gunning, David: XAI—Explainable Artificial Intelligence. Science Robots (2019)

\bibitem{hein2019relu}
Hein, Matthias and Andriushchenko, Maksym and Bitterwolf, Julian: Why Relu Networks Yield High-Confidence Predictions Far Away From the Training Data and How to Mitigate the Problem. CVPR (2019)

\bibitem{howard2019searching}
Howard, Andrew and Sandler, Mark and Chu, Grace and Chen, Liang-Chieh and Chen, Bo and Tan, Mingxing and Wang, Weijun and Zhu, Yukun and Pang, Ruoming and Vasudevan, Vijay and others: Searching for Mobilenetv3. ICCV (2019)

\bibitem{is}
Tim Salimans and Ian Goodfellow and Wojciech Zaremba and Vicki Cheung and Alec Radford and Xi Chen: Improved Techniques for Training GANs. NeurIPS (2016)

\bibitem{kenny2021Plausible}
Kenny, Eoin M. and Keane, Mark T.: On {{Generating Plausible Counterfactual}} and {{Semi-Factual Explanations}} for {{Deep Learning}}. AAAI (2021)

\bibitem{lage2019evaluation}
Isaac Lage and Emily Chen and Jeffrey He and Menaka Narayanan and Been Kim and Sam Gershman and Finale Doshi-Velez: An Evaluation of the Human-Interpretability of Explanation. NeurIPS (2018)

\bibitem{macem}
Amit Dhurandhar and Tejaswini Pedapati and Avinash Balakrishnan and Pin-Yu Chen and Karthikeyan Shanmugam and Ruchir Puri: Model Agnostic Contrastive Explanations for Structured Data. arXiv:1906.00117 (2019)

\bibitem{mnist}
LeCun, Yann and Bottou, L{\'e}on and Bengio, Yoshua and Haffner, Patrick: Gradient-Based Learning Applied to Document Recognition. IEEE (1998)

\bibitem{pcattgan}
Barredo-Arrieta, Alejandro and Del Ser, Javier: Plausible Counterfactuals: Auditing Deep Learning Classifiers With Realistic Adversarial Examples. IJCNN (2020)

\bibitem{piece}
Kenny, Eoin M and Keane, Mark T: On Generating Plausible Counterfactual and Semi-Factual Explanations for Deep Learning. AAAI (2021)

\bibitem{precrecall}
Sajjadi, Mehdi SM and Bachem, Olivier and Lucic, Mario and Bousquet, Olivier and Gelly, Sylvain: Assessing Generative Models via Precision and Recall. NeurIPS (2018)

\bibitem{revise}
Shalmali Joshi and Oluwasanmi Koyejo and Warut Vijitbenjaronk and Been Kim and Joydeep Ghosh: Towards Realistic Individual Recourse and Actionable Explanations in Black-Box Decision Making Systems. arXiv:1907.09615 (2019)

\bibitem{rosenfeld2021better}
Rosenfeld, Avi: Better Metrics for Evaluating Explainable Artificial Intelligence. AAMAS (2021)

\bibitem{rw_diff1}
Jeanneret, Guillaume and Simon, Lo{\"\i}c and Jurie, Fr{\'e}d{\'e}ric: Diffusion Models for Counterfactual Explanations. ACCV (2022)

\bibitem{sahakyanExplainableArtificialIntelligence2021}
Sahakyan, Maria and Aung, Zeyar and Rahwan, Talal: Explainable {{Artificial Intelligence}} for {{Tabular Data}}: {{A Survey}}. IEEE Access (2021)

\bibitem{salimans2017pixelcnn++}
Tim Salimans and Andrej Karpathy and Xi Chen and Diederik P. Kingma: PixelCNN++: Improving the PixelCNN With Discretized Logistic Mixture Likelihood and Other Modifications. ICLR (2017)

\bibitem{song2020score}
Yang Song and Jascha Sohl-Dickstein and Diederik P. Kingma and Abhishek Kumar and Stefano Ermon and Ben Poole: Score-Based Generative Modeling Through Stochastic Differential Equations. ICLR (2021)

\bibitem{theis2015note}
Lucas Theis and Aäron van den Oord and Matthias Bethge: A Note on the Evaluation of Generative Models. ICLR (2016)

\bibitem{tomczak2018vae}
Tomczak, Jakub and Welling, Max: VAE With a VampPrior. AISTATS (2018)

\bibitem{understandingdiffusion}
Calvin Luo: Understanding Diffusion Models: {A} Unified Perspective. arXiv:2208.11970 (2022)

\bibitem{Vaeth2023}
Philipp V{\"{a}}th and
Alexander M. Fr{\"{u}}hwald and
Benjamin Paa{\ss}en and
Magda Gregorova: Diffusion-Based Visual Counterfactual Explanations - Towards Systematic
Quantitative Evaluation. ECML PKDD XKDD Workshop (2023)

\bibitem{visual_counterfactuals}
Goyal, Yash and Wu, Ziyan and Ernst, Jan and Batra, Dhruv and Parikh, Devi and Lee, Stefan: Counterfactual Visual Explanations. ICML (2019)

\bibitem{von-platen-etal-2022-diffusers}
Patrick von Platen and Suraj Patil and Anton Lozhkov and Pedro Cuenca and Nathan Lambert and Kashif Rasul and Mishig Davaadorj and Thomas Wolf: Diffusers: State-of-the-Art Diffusion Models. GitHub (2022)

\bibitem{xaisurvey_1}
Adadi, Amina and Berrada, Mohammed: Peeking Inside the Black-Box: A Survey on Explainable Artificial Intelligence (XAI). IEEE access (2018)

\bibitem{xaisurvey_2}
Arun Das and Paul Rad: Opportunities and Challenges in Explainable Artificial Intelligence (XAI): A Survey. arXiv:2006.11371 (2020)

\bibitem{xaisurvey_3}
Sahil Verma and Varich Boonsanong and Minh Hoang and Keegan E. Hines and John P. Dickerson and Chirag Shah: Counterfactual Explanations and Algorithmic Recourses for Machine Learning: A Review. ACM Comput. Surv. (2022)

\bibitem{yangDiffusionModelsComprehensive2023}
Yang, Ling and Zhang, Zhilong and Song, Yang and Hong, Shenda and Xu, Runsheng and Zhao, Yue and Zhang, Wentao and Cui, Bin and Yang, Ming-Hsuan: Diffusion {{Models}}: {{A Comprehensive Survey}} of {{Methods}} and {{Applications}}. ACM Comput. Surv. (2023)

\end{thebibliography}
